\documentclass{article}

\usepackage{arxiv}

\usepackage[utf8]{inputenc} 
\usepackage[T1]{fontenc}    
\usepackage{hyperref}       
\usepackage{url}            
\usepackage{booktabs}       
\usepackage{amsfonts}       
\usepackage{nicefrac}       
\usepackage{microtype}      
\usepackage{cleveref}       
\usepackage{lipsum}         
\usepackage{graphicx}
\usepackage[numbers]{natbib}
\usepackage{doi}
\usepackage{lscape} 

\title{
	A simple DNN regression for the chemical composition in essential oil 
}

\newif\ifuniqueAffiliation

\uniqueAffiliationtrue
\ifuniqueAffiliation 
\author{\href{https://orcid.org/0009-0000-4254-7803}{\includegraphics[scale=0.06]{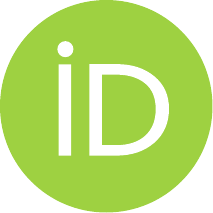}\hspace{1mm}Yuki Harada} \\
		Priority Organization for Innovation and Excellence Laboratory for Data Sciences, \\
		Kumamoto University, Kurokami 2-39-1, Chuo-ku, Kumamoto 860-8555, JP \\
	\texttt{yharada@kumamoto-u.ac.jp} \\
	\And
	\href{}{\hspace{1mm}Shuichi Maeda} \\
		Priority Organization for Innovation and Excellence Laboratory for Data Sciences, \\
		Kumamoto University, Kurokami 2-39-1, Chuo-ku, Kumamoto 860-8555, JP \\
	\texttt{Shuichi2503736@gmail.com} \\
	\And
	\href{https://orcid.org/0000-0003-0094-8179}{\includegraphics[scale=0.06]{fig/orcid.pdf}\hspace{1mm}Masato Kiyama} \\
		Faculty of Advanced Science and Technology, \\
		Kumamoto University, Kurokami 2-39-1, Chuo-ku, Kumamoto 860-8555, JP \\
	\texttt{masato@cs.kumamoto-u.ac.jp} \\
	\And
	\href{https://orcid.org/0000-0002-6437-6993}{\includegraphics[scale=0.06]{fig/orcid.pdf}\hspace{1mm}Shinichiro Nakamura} \\
		Priority Organization for Innovation and Excellence Laboratory for Data Sciences, \\
		Kumamoto University, Kurokami 2-39-1, Chuo-ku, Kumamoto 860-8555, JP \\
	\texttt{shindon@kumamoto-u.ac.jp} \\
}

\else
\usepackage{authblk}

\newbox{\orcid}\sbox{\orcid}{\includegraphics[scale=0.06]{fig/orcid.pdf}} 
\author[1]{%
	\href{https://orcid.org/0009-0000-4254-7803}{\usebox{\orcid}\hspace{1mm}Yuki Harada\thanks{\texttt{yharada@kumamoto-u.ac.jp}}}
}
\author[1]{%
	\href{https://orcid.org/}{\usebox{\orcid}\hspace{1mm}Shuichi Maeda\thanks{\texttt{Shuichi2503736@gmail.com}}}
}
\author[4]{%
	\href{https://orcid.org/0000-0003-0094-8179}{\usebox{\orcid}\hspace{1mm}Masato Kiyama\thanks{\texttt{masato@cs.kumamoto-u.ac.jp}}}
}
\author[1]{%
	\href{https://orcid.org/0000-0002-6437-6993}{\usebox{\orcid}\hspace{1mm}Shinichiro Nakamura\thanks{\texttt{shindon@kumamoto-u.ac.jp}}}
}
\affil[1]{ \\
	Priority Organization for Innovation and Excellence Laboratory for Data Sciences, \\
	Kumamoto University, Kurokami 2-39-1, Chuo-ku, Kumamoto 860-8555, JP \\
}
\affil[4]{ \\
	Faculty of Advanced Science and Technology, \\
	Kumamoto University, Kurokami 2-39-1, Chuo-ku, Kumamoto 860-8555, JP \\
}
\fi

\renewcommand{\shorttitle}{
	A simple DNN regression for the chemical composition in essential oil 
}

\hypersetup{
pdftitle={
	A simple DNN regression for the chemical composition in essential oil 
},
pdfsubject={
	化合物組成をもとにした精油情報の予測のための深層学習構成の検討
},
pdfauthor={
	Yuki Harada, 
	Shuichi Maeda, 
	Masato Kiyama, 
	Shinichiro Nakamura
},
pdfkeywords={essential oil, chemical composition, chemical space},
}

\begin{document}
\maketitle

\begin{abstract}
Although experimental design and methodological surveys for mono-molecular activity/property has been extensively investigated, 
those for chemical composition have received little attention, with the exception of a few prior studies. 
In this study, we configured three simple DNN regressors to predict essential oil property based on chemical composition. 
Despite showing overfitting due to the small size of dataset, all models were trained effectively in this study. 

\end{abstract}


\section{Introduction}

Experimental design and methodological survey for the mono-molecular activity/property, 
including in vitro and in vivo bioassays, in-silico screening from chemical databases based on similarity measures, and quantitative structure-activity relationship (QSAR) investigations on lead molecules, 
has been extensively investigated.
\citep{keller2017predicting, neves2018qsar}
In other hand, 
experimental design and methodological survey for the activity/property of chemical composition, 
in which a mixture of molecules is expected to produce function or activity, 
has been paid little attention.

This study focused on analytical tables of essential oils and emploied a convolutional neural network (CNN) \citep{zhang2015sensitivity}
and two types of Graph Neural Network (GNN(s); GCNconv \citep{kipf2016semi, pyggcnconv} and GATconv \citep{velivckovic2017graph, pyggatconv} ) as a regressor to predict their properties based on composition. 
We build a simple deep neural network (DNN) regressor 
for essential oil properties based on chemical composition and how it works.

%
%
%

\subsection{Chemical space of plant metabolome / semio-chemicals}

The essential oils and their molecular component has a long history with the aromatherapy. 
They have been employed as a type of traditional or alternative medicine for the management of various human diseases.
This complementary therapy is thought to improve psychological well-being up to the present.

A semiochemical is a mixture or single molecule that carries a message for communication purposes, such as pheromones and defense mechanisms that allow animals to recognize and detect individuals, in Behavioral Biology and Ecology sentense. 
Over the past decades, scientific investigations have revealed that volatile compounds; with a molecular weight <300 and high vapor pressure, play a key role in life by acting as semiochemicals, mediating inter- and intraspecies interactions of living organisms.
Their molecular component is stemming from major pathways of secondary metabolism in plant, and that potentially acts as semiochemical(s). 
\citep{dunkel2009superscent, pickett2016plant}

There are many cases in which the responsible molecule has been identified for the communication, 
but there are also many cases where it has not. 
The reasons for this difficulty include the possibility that the some communication might be caused by a combination of molecules. 
This research will provide an analysis of such responsibilities.

%
%
%
%

\section{Experiments}
\subsection{Essential oil database }

The property table for essential oils is collected from a website and reshaped, \citep{kumar2018aromadb} 
as shown in Fig.1.
The property table includes 'Essential Oil Name', 'Plant Name', 'Plant Tissue Name', 'Compound Name(s)' (chemical names only) and detailed web link. 
By the detailed web link in the property table, 
we can obtain the analytical table including 'Compound Name(s)' and 'Compound Percentage(\%)'. 
The area\% of gas chromatography (GC) is denoted as 'Compound Percentage(\%)' in the analytical table. 
The molecular structure is able to obtain from 'Compound Name(s)'.

\begin{figure} 
	\centering
	\includegraphics[keepaspectratio, width=6.5 in ]{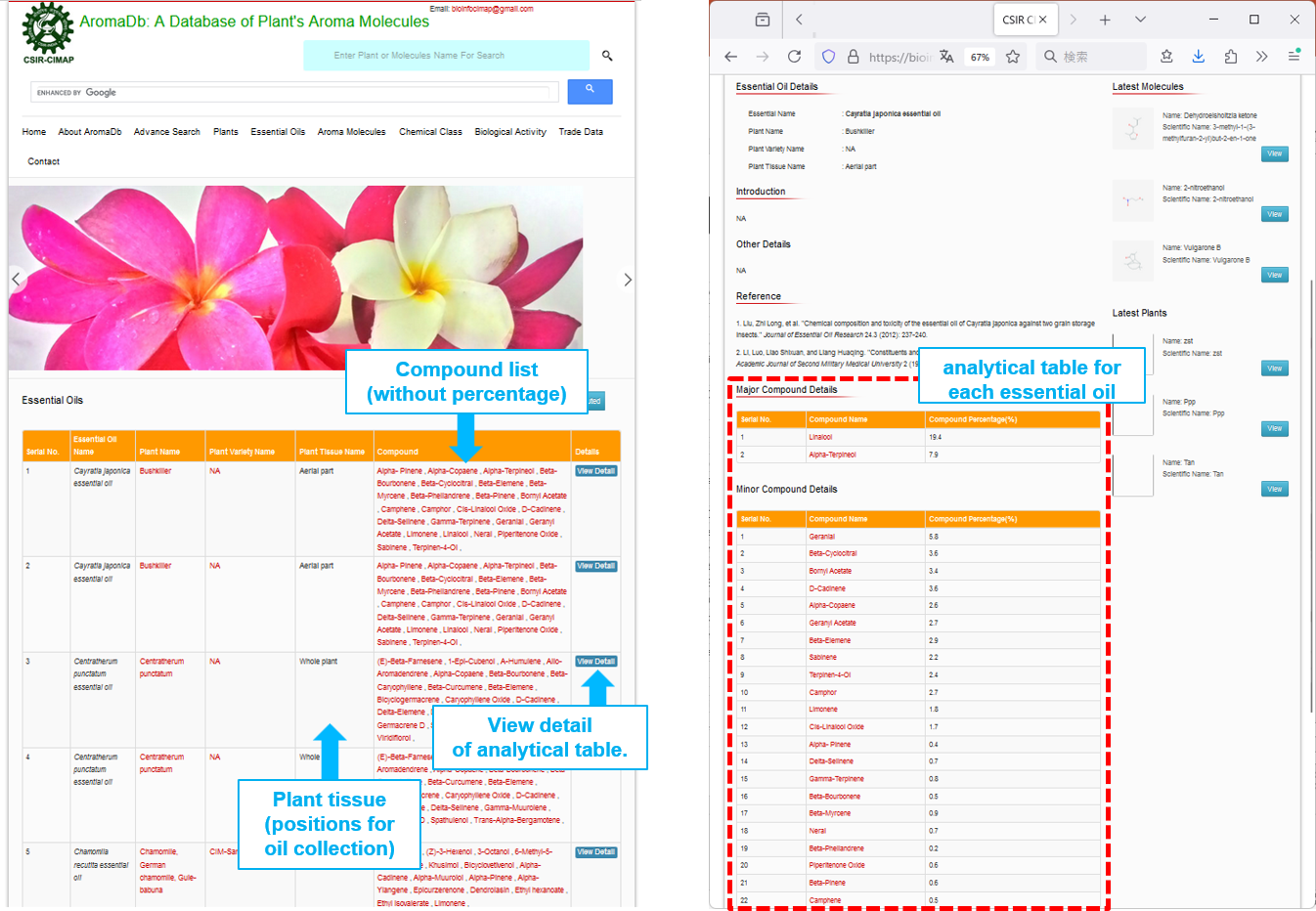}
	\caption{Data in website for the property table and analytical table }
\end{figure} 

%

\subsection{DNN input / output preparation and the evaluation of predictability }

In PyTorch, preparing a DataLoader or Dataset is essential for efficiently and effectively training deep neural networks (and also validating). 
It is crucial for efficiently managing and processing data. 
They provide flexibility, efficiency, and ease of use, which are essential for training deep neural networks effectively.

The dataloaders for CNN input data ('Input Stucked Node Data') and GNN input data ('Input Node Data' and 'Input Edge Data') 
are prepared based on both of the property table and the analytical table (see also Fig.1).  
There are several possible design for input data in DNN, in this study, we designed it as follows.
'Compound Percentage(area\%)' is approximately proportional to the weight content ratio of the component. 
Some area\% value is displayed as 'Trace', we filled with small value of 0.01. 


'Input Node Data' is a 1D tensor that concatenate of the area\% and the molecular fingerprint, 
where following four fingerprint are abailable: MACCS keys\citep{durant2002reoptimization}, ECFPs \citep{rogers2010extended}, Avalon fingerprints \citep{gedeck2006qsar}, and RD-kit fingerprints\citep{wwwrdkit}. 
'Input Stucked Node Data' is a 2D tensor obtained by stacking of 'Input Node Data'. 
The index numbers in CNN input data and the node numbers in GNN input data represent the number of chemical compounds, 
whose size varies depending on the number of chemicals in essential oil. 
The GNN model is to be a complete graph, that assumed 'Input Edge Data' between all nodes.
'Input Edge Data' is to be a similarity measure based on fingerprints.

The target data, which corresponds to the output of a DNN model, was designed as multi-target in this study. 
Instead of the categorial 'Plant Tissue Name' in the property table for essential oils, 
the entry is encorded into a multilabeled binary table. 
The multilabeled target was based on nine categories, dropping those having fewer than five data points.
Fig.2 depicts a bar graph representing the number of data points corresponding to 'Plant Tissue Name' in the property table.

\begin{figure} 
	\centering
	\includegraphics[keepaspectratio, width=4.0 in ]{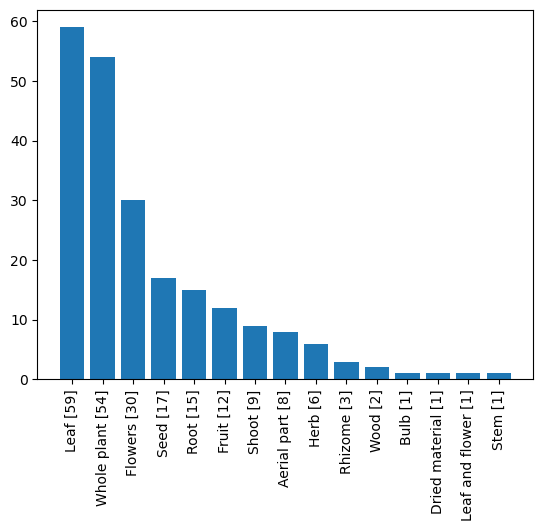}
	\caption{Data points of each 'Plant Tissue Name' in the property table}
\end{figure} 

We utilized K-fold CV in Sklearn to evaluate the DNN predictability through cross-validation. 
AUC was calculated for training epoch numbers for epoch target.

%
%
%

\subsection{Three architecture and two loss design }

We used three architectures for the deep learning model in this study: CNN \citep{zhang2015sensitivity}, GCNconv \citep{kipf2016semi, pyggcnconv}, and GATconv \citep{velivckovic2017graph, pyggatconv}. 
We also used two different options for the final activation function, model output (objective variable in DNN), and loss function as shown in Table 1.

\begin{table} 
    \centering
    \caption{Two Loss Design }
    \begin{tabular}{|l|c|c|p{3.0 in}|} \hline
		loss function &  (suitable-final) activation function & model output \\ \hline
		BCEWithLogitsLoss & Linear & OneHotTarget(Normal) \\ \hline
		NLL\_loss & Log\_SoftMax & OneHotTarget(Normal+Inversed) \\ \hline
	\end{tabular}
\end{table} 

%
%

\section{Disuccusion}

Three architectures and two loss designs are emploied in this investigation. 
Fig.3 displays the AUC history for each condition. 
The horizontal axis represents the number of training epochs, whereas the vertical axis represents the test loss. 
Despite showing overfitting due to the small size of dataset, all models were trained effectively. 
Table 2 summarizes the AUCs in this investigation.

\begin{figure} 
	\centering
	\includegraphics[keepaspectratio, width=6.5 in ]{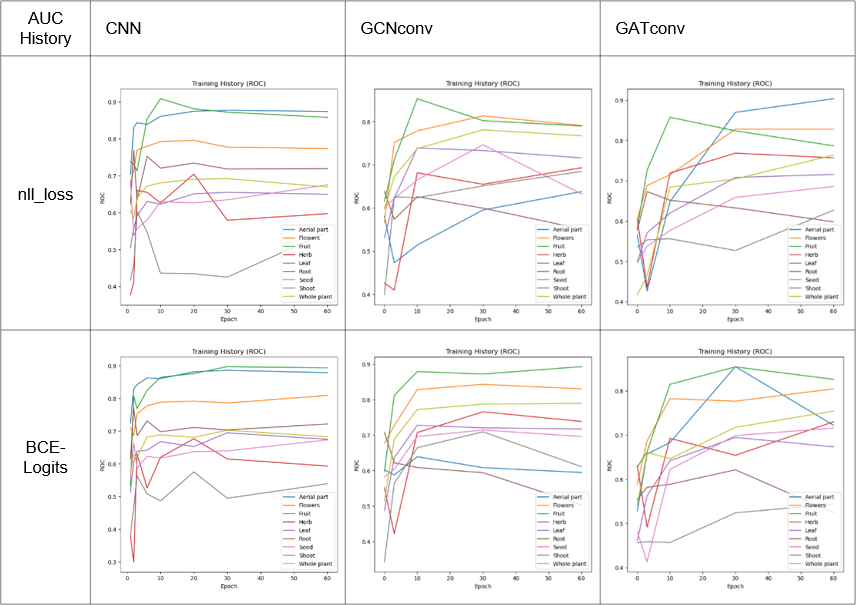}
	\caption{AUC history by three regressors with two loss designs; (i) CNN, (ii) GCNconv, and (iii) GATconv }
\end{figure} 

\begin{table} 
	\centering
	\includegraphics[keepaspectratio, width=6.5 in ]{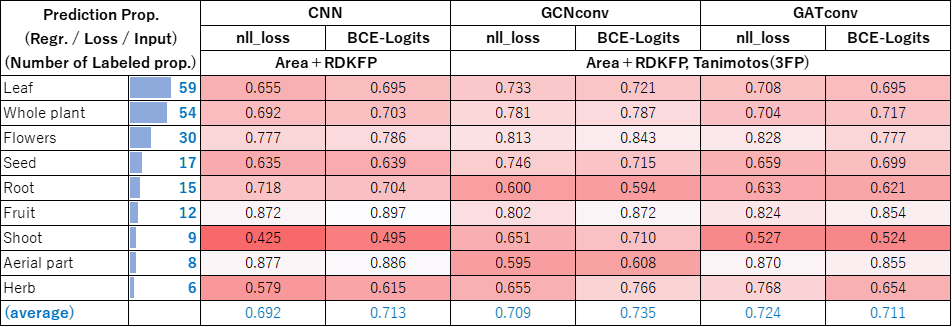}
	\caption{The AUCs by three regressors with two loss designs in epoch number = 30; (i) CNN, (ii) GCNconv, and (iii) GATconv }
\end{table} 

The fourth entry (GCNconv with BCEWithLogitsLoss) and fifth entry (GATconv with NLL\_loss) demonstrate good predictability. 
The ROC curves for each target in the prediction by GATconv with NLL\_loss is presented in Fig.4.

\begin{figure} 
	\centering
	\includegraphics[keepaspectratio, width=6.5 in ]{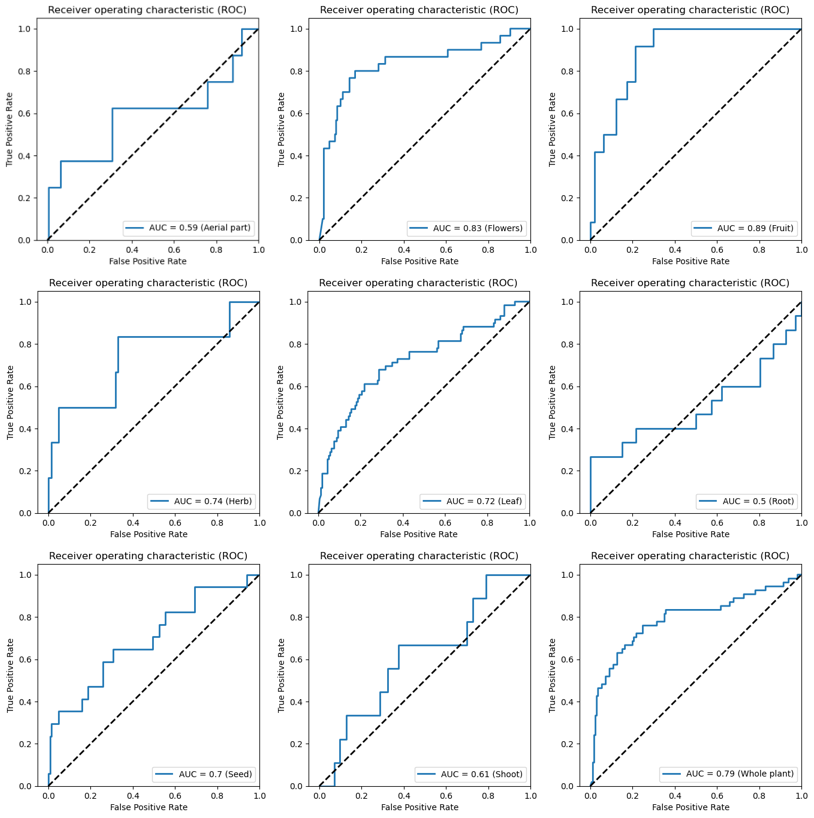}
	\caption{ROC plot in the prediction by GATconv with NLL\_loss (5th entry in Table 2) }
\end{figure} 

%
%

\subsection{Comparison with related research}

We will disuccuss the difference between our regression and the related previous research by sisson et.al. \citep{sisson2023olfactory, GitHubOdorPair}. 
Sisson et.al. boldly designed the whole chemical space as a perspective using graph neural network technology, which is an innovative idea. 
In both of Sisson's work and this study, the node designs are currently based on molecular structural representation (molecular structural fingerprints).
Sisson et al.'s neural networks focus on designing the whole chemical space as a large hypergraph. 
In this hypergraph, each node corresponds to a molecule, which has a representation connected to the surrounding nodes and edges.  
%
%
Our DNNs are designed to process a chemical composition as a graph and output the essential oil properties. 
Although there is no design for the whole chemical space in our current DNN, 
it may be necessary to consider vector representation other than the molecular structure itself.



\section{Future task}
\subsection{Output target of DNN }

Sensory evaluation of essential oils: While prediction of odor character may be of interest, there is currently no database available in sensory evaluation of essential oil. 
The availablity of sensory evaluation differs from the task on a task of single odorant, 
for which some databases are available to study the predictability of sensory evaluation. \citep{castro2022pyrfume, githubpyrfume} 
In this research, we did not forecast the odor type of essential oils using our architecture. 
If any database with odor type information is available, we will evaluate the predictability on our architecture.

The scientific name and/or varieties of the plant: The scientific name can be obtained via the web link of the entry "Detail" in the essential oil property table. 
However, they were unsuitable for learning and verification due to their diversity.
We attempted to compare plant varieties in the essential oil property table using the POWO database \citep{powo_plants_2024}, 
but were unable to do so due to inconsistencies in the scientific names and lack of IDs.

%
%
%

\section{Conclusion}

For deep learning, the design of input and output data, as well as architecture selection, are critical. 
In this study, a simple three DNN regressor for essential oil property based on chemical composition was presented. 
It used a molecular fingerprint, an essential oil attribute ('Plant Tissue Name'), and a GNN or CNN as inputs, outputs, and architecture. 
Despite showing overfitting due to the small size of dataset, all models were trained effectively.

\section{Acknowledgments}

The present work is supported by The Chemo-Sero-Therapeutic Research Institute (KAKETSUKEN), Japan. 
We would like to express our sincere gratitude to Associate Professor Mikiyo Wada, 
of Kumamoto University, for notifying us about the existence of the POWO and some plants databases. 
We are also grateful to the Aroma Environment Association of Japan (AEAJ), 
whose database was not directly involved in this study but whose activity and oil exhibition provided a lot of inspiration.
The source code and related materials are available at our GitHub repository. \citep{ourrepo_github} 

\bibliographystyle{unsrtnat}
\bibliography{references}  

\begin{thebibliography}{20}
\providecommand{\natexlab}[1]{#1}
\providecommand{\url}[1]{\texttt{#1}}
\expandafter\ifx\csname urlstyle\endcsname\relax
  \providecommand{\doi}[1]{doi: #1}\else
  \providecommand{\doi}{doi: \begingroup \urlstyle{rm}\Url}\fi

\bibitem[Keller et~al.(2017)Keller, Gerkin, Guan, Dhurandhar, Turu, Szalai, Mainland, Ihara, Yu, Wolfinger, et~al.]{keller2017predicting}
Andreas Keller, Richard~C Gerkin, Yuanfang Guan, Amit Dhurandhar, Gabor Turu, Bence Szalai, Joel~D Mainland, Yusuke Ihara, Chung~Wen Yu, Russ Wolfinger, et~al.
\newblock Predicting human olfactory perception from chemical features of odor molecules.
\newblock \emph{Science}, 355\penalty0 (6327):\penalty0 820--826, 2017.

\bibitem[Neves et~al.(2018)Neves, Braga, Melo-Filho, Moreira-Filho, Muratov, and Andrade]{neves2018qsar}
Bruno~J Neves, Rodolpho~C Braga, Cleber~C Melo-Filho, Jos{\'e}~Te{\'o}filo Moreira-Filho, Eugene~N Muratov, and Carolina~Horta Andrade.
\newblock Qsar-based virtual screening: advances and applications in drug discovery.
\newblock \emph{Frontiers in pharmacology}, 9:\penalty0 1275, 2018.

\bibitem[Zhang and Wallace(2015)]{zhang2015sensitivity}
Ye~Zhang and Byron Wallace.
\newblock A sensitivity analysis of (and practitioners' guide to) convolutional neural networks for sentence classification.
\newblock \emph{arXiv preprint arXiv:1510.03820}, 2015.

\bibitem[Kipf and Welling(2016)]{kipf2016semi}
Thomas~N Kipf and Max Welling.
\newblock Semi-supervised classification with graph convolutional networks.
\newblock \emph{arXiv preprint arXiv:1609.02907}, 2016.

\bibitem[GCNconv@torchgeometric()]{pyggcnconv}
GCNconv@torchgeometric.
\newblock \url{https://github.com/pyg-team/pytorch_geometric/blob/master/torch_geometric/nn/conv/gcn_conv.py}.

\bibitem[Veli{\v{c}}kovi{\'c} et~al.(2017)Veli{\v{c}}kovi{\'c}, Cucurull, Casanova, Romero, Lio, and Bengio]{velivckovic2017graph}
Petar Veli{\v{c}}kovi{\'c}, Guillem Cucurull, Arantxa Casanova, Adriana Romero, Pietro Lio, and Yoshua Bengio.
\newblock Graph attention networks.
\newblock \emph{arXiv preprint arXiv:1710.10903}, 2017.

\bibitem[GATconv@torchgeometric()]{pyggatconv}
GATconv@torchgeometric.
\newblock \url{https://github.com/pyg-team/pytorch_geometric/blob/master/torch_geometric/nn/conv/gat_conv.py}.

\bibitem[Dunkel et~al.(2009)Dunkel, Schmidt, Struck, Berger, Gruening, Hossbach, Jaeger, Effmert, Piechulla, Eriksson, et~al.]{dunkel2009superscent}
Mathias Dunkel, Ulrike Schmidt, Swantje Struck, Lena Berger, Bjoern Gruening, Julia Hossbach, Ines~S Jaeger, Uta Effmert, Birgit Piechulla, Roger Eriksson, et~al.
\newblock Superscent—a database of flavors and scents.
\newblock \emph{Nucleic acids research}, 37\penalty0 (suppl\_1):\penalty0 D291--D294, 2009.

\bibitem[Pickett and Khan(2016)]{pickett2016plant}
John~A Pickett and Zeyaur~R Khan.
\newblock Plant volatile-mediated signalling and its application in agriculture: successes and challenges.
\newblock \emph{New Phytologist}, 212\penalty0 (4):\penalty0 856--870, 2016.

\bibitem[Kumar et~al.(2018)Kumar, Prakash, Tripathi, Tandon, Gupta, Rahman, Lal, Semwal, Darokar, and Khan]{kumar2018aromadb}
Yogesh Kumar, Om~Prakash, Himanshu Tripathi, Sudeep Tandon, Madan~M Gupta, Laiq-Ur Rahman, Raj~K Lal, Manoj Semwal, Mahendra~Pandurang Darokar, and Feroz Khan.
\newblock Aromadb: a database of medicinal and aromatic plant’s aroma molecules with phytochemistry and therapeutic potentials.
\newblock \emph{Frontiers in plant science}, 9:\penalty0 1081, 2018.

\bibitem[Durant et~al.(2002)Durant, Leland, Henry, and Nourse]{durant2002reoptimization}
Joseph~L Durant, Burton~A Leland, Douglas~R Henry, and James~G Nourse.
\newblock Reoptimization of mdl keys for use in drug discovery.
\newblock \emph{Journal of chemical information and computer sciences}, 42\penalty0 (6):\penalty0 1273--1280, 2002.

\bibitem[Rogers and Hahn(2010)]{rogers2010extended}
David Rogers and Mathew Hahn.
\newblock Extended-connectivity fingerprints.
\newblock \emph{Journal of chemical information and modeling}, 50\penalty0 (5):\penalty0 742--754, 2010.

\bibitem[Gedeck et~al.(2006)Gedeck, Rohde, and Bartels]{gedeck2006qsar}
Peter Gedeck, Bernhard Rohde, and Christian Bartels.
\newblock Qsar- how good is it in practice? comparison of descriptor sets on an unbiased cross section of corporate data sets.
\newblock \emph{Journal of chemical information and modeling}, 46\penalty0 (5):\penalty0 1924--1936, 2006.

\bibitem[cheminformatics()]{wwwrdkit}
RDKit: Open-source cheminformatics.
\newblock \url{https://www.rdkit.org}.

\bibitem[Sisson et~al.(2023)Sisson, Barsainyan, Sharma, and Kumar]{sisson2023olfactory}
Laura Sisson, Aryan~Amit Barsainyan, Mrityunjay Sharma, and Ritesh Kumar.
\newblock Olfactory label prediction on aroma-chemical pairs.
\newblock \emph{arXiv preprint arXiv:2312.16124}, 2023.

\bibitem[et. al.()]{GitHubOdorPair}
Laura~Sisson et. al.
\newblock Github - odor-pair.
\newblock \url{https://github.com/odor-pair/odor-pair}.

\bibitem[Castro et~al.(2022)Castro, Gould, Pellegrino, Liang, Coleman, Patel, Wallace, Bhatnagar, Mainland, and Gerkin]{castro2022pyrfume}
Jason~B Castro, Travis~J Gould, Robert Pellegrino, Zhiwei Liang, Liyah~A Coleman, Famesh Patel, Derek~S Wallace, Tanushri Bhatnagar, Joel~D Mainland, and Richard~C Gerkin.
\newblock Pyrfume: A window to the world’s olfactory data.
\newblock \emph{bioRxiv}, pages 2022--09, 2022.

\bibitem[Project()]{githubpyrfume}
The~Pyrfume Project.
\newblock \url{https://github.com/pyrfume}.

\bibitem[{POWO}(2024)]{powo_plants_2024}
{POWO}.
\newblock Plants of the {World} {Online}, 2024.
\newblock URL \url{http://www.plantsoftheworldonline.org/}.

\bibitem[our(2024)]{ourrepo_github}
Chemical composition regression.
\newblock \url{https://github.com/yhua0917/ChemicalCompositionRegression/}, 2024.

\end{thebibliography}






\end{document}